\documentclass[sigconf]{acmart}

\usepackage{multirow}
\usepackage{url}
\usepackage{color, colortbl}
\usepackage{soul}
\AtBeginDocument{%
  }

\copyrightyear{2026}
\acmYear{2026}
\setcopyright{cc}
\setcctype{by}
\acmConference[MM '26]{Proceedings of the 34th ACM International Conference on Multimedia}
  {November 10--14, 2026}{Rio de Janeiro, Brazil}
\acmBooktitle{Proceedings of the 34th ACM International Conference on Multimedia
  (MM '26), November 10--14, 2026, Rio de Janeiro, Brazil}
\acmDOI{10.1145/3767308.3835205}
\acmISBN{979-8-4007-2213-4/2026/11}
\acmSubmissionID{1764}

\begin{document}
\title{HarMoE: Multi-Source Chest Radiograph Pretraining with Dataset-Disentangled Experts}

\author{Haozhe Luo}
\authornote{Haozhe Luo and Ziyu Zhou contributed equally to this work.}
\authornote{Corresponding author.}
\orcid{0009-0009-8021-2073}
\email{haozhe.luo@unibe.ch}
\affiliation{%
  \institution{ARTORG Center for Biomedical Engineering Research, University of Bern}
  \city{Bern}
  \country{Switzerland}
}
\affiliation{%
  \institution{Kaiko.AI}
  \city{Zurich}
  \country{Switzerland}
}

\author{Ziyu Zhou}
\authornotemark[1]
\orcid{0009-0000-9805-3546}
\email{zhouziyu@sjtu.edu.cn}
\affiliation{%
  \institution{Shanghai Jiao Tong University}
  \city{Shanghai}
  \country{China}
}

\author{Shelley Zixin Shu}
\orcid{0009-0008-0392-7924}
\email{zixin.shu@unibe.ch}
\affiliation{%
  \institution{ARTORG Center for Biomedical Engineering Research, University of Bern}
  \city{Bern}
  \country{Switzerland}
}

\author{Mauricio Reyes}
\orcid{0000-0002-2434-9990}
\email{mauricio.reyes3@unibe.ch}
\affiliation{%
  \institution{ARTORG Center for Biomedical Engineering Research, University of Bern}
  \city{Bern}
  \country{Switzerland}
}

\renewcommand{\shortauthors}{Haozhe Luo, Ziyu Zhou, Shelley Zixin Shu, and Mauricio Reyes}

\begin{abstract}
Recent vision-language models for chest X-ray understanding are largely built on image-report alignment and therefore rely heavily on MIMIC-CXR as the dominant pretraining source. While effective at scale, this paradigm underexplores an important alternative source of supervision: a range of existing multi-label classification datasets, which provide cleaner and more explicit disease signals than free-text reports, and can offer broader pathology coverage when combined across sources. However, learning from such heterogeneous datasets is nontrivial, as differences in label ontologies, annotation protocols, acquisition pipelines, and report styles can cause models to entangle clinical semantics with dataset identity, leading to poor transfer despite increased scale. In this work, we revisit radiology VLM construction from the perspective of harmonized multi-source learning. We propose HarMoE, a dataset-aware mixture-of-experts framework that learns shared cross-dataset medical semantics while confining source-specific variation to lightweight residual experts in deeper decoder layers. To further exploit clean supervision from labeled datasets, we train in a unified disease vocabulary with masked multi-dataset supervision, enabling the model to leverage complementary annotations without introducing false negatives. Experiments on large-scale chest X-ray benchmarks show that HarMoE consistently improves zero-shot classification, out-of-distribution transfer, and grounding over strong baselines. Our results suggest that building robust radiology VLMs requires moving beyond single-source image-report alignment toward structured knowledge construction from heterogeneous datasets with cleaner supervision and broader coverage. Code and the 873k harmonized dataset will be released at https://github.com/Roypic/harmoe.

\end{abstract}


\maketitle

\section{Introduction}

\begin{figure}[h]
    \centering
    \includegraphics[width=\linewidth]{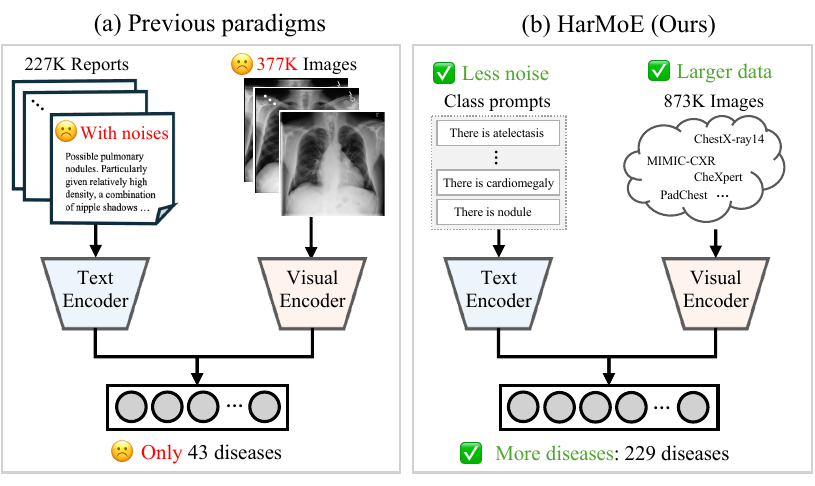}
    \Description{A two-panel schematic contrasts single-source image-report pretraining with HarMoE multi-source pretraining over four chest X-ray datasets and a unified 229-class disease vocabulary.}
    \caption{Single-source vs.\ multi-source pretraining for chest X-ray understanding. (a) Existing VLP methods depend on paired image-report data from a single source (e.g., MIMIC-CXR: 377K images, 43 conditions). (b) HarMoE harmonizes four heterogeneous datasets into a unified 229-class vocabulary, scaling to 873K images while accommodating sources with and without free-text reports.}
    \label{fig:motivation}
\end{figure}

Chest radiography is among the most frequently performed diagnostic 
imaging procedures worldwide, yet its interpretation remains 
labor-intensive and subject to substantial inter-observer 
variability~\cite{gundel2021robust,clark1995interobserver}. This 
bottleneck has driven sustained interest in vision-language 
pretraining (VLP): by aligning chest X-ray images with radiology 
reports through contrastive learning, methods such as 
ConVIRT~\cite{zhang2022contrastive}, GLoRIA~\cite{huang2021gloria}, 
KAD~\cite{zhang2023knowledge}, and MedKLIP~\cite{wu2023medklip} 
enable zero-shot classification without manual annotation. More 
recently, CarZero~\cite{lai2024carzero} and 
RadZero~\cite{park2025radzero} have shown that classification-level 
prompts yield cleaner supervision while retaining semantic grounding. 
Despite this diversity, the field remains constrained by a shared 
dependency on a single institutional source, predominantly 
MIMIC-CXR~\cite{johnson2019mimic} (377K images, 43 conditions) as shown in Fig. \ref{fig:motivation} (a).

This single-source dependency is limiting because data scale and 
diversity are among the strongest predictors of representation 
quality. CLIP~\cite{radford2021learning} and ALIGN~\cite{jia2021scaling} 
demonstrated robust zero-shot transfer from massive image-text 
corpora, with subsequent work confirming power-law scaling with 
data size~\cite{cherti2023reproducible}. Radiology cannot replicate 
this directly: datasets are institutionally siloed, and the majority 
of large-scale chest X-ray collections discard free-text reports in 
favor of structured labels. ChestX-ray14~\cite{wang2017chestx} 
(112K), CheXpert~\cite{irvin2019chexpert} (224K), and 
PadChest~\cite{bustos2020padchest} (160K) are inaccessible to 
standard contrastive VLP pipelines despite their clinical value. 
Combining them with report-bearing corpora would more than double 
the pretraining pool, from 377K to 873K images, while expanding 
pathology coverage from 43 to 229 conditions (Fig.~\ref{fig:motivation} (b)).

The technical barrier to such aggregation is not data access but \emph{representation 
quality under multi-source supervision}. Each dataset reflects a 
distinct combination of acquisition hardware, patient demographics, 
and annotation protocol~\cite{oakden2020hidden, zech2018variable}; 
naively pooling sources induces spurious correlations between 
dataset identity and diagnostic labels~\cite{degrave2021ai}, and 
treating unannotated classes as negative introduces systematic 
false negatives that distort the learned decision 
boundary~\cite{bekker2020learning}. These failure modes are 
well-studied in multi-domain learning~\cite{geirhos2020shortcut}, 
but existing VLP methods do not address them directly, as they 
operate within single-source pipelines where they do not arise 
by construction.

We approach this as a \textbf{multi-source supervised pretraining 
problem with partial labels and heterogeneous domain shifts}, and 
hypothesize that effective multi-source learning requires 
architecturally separating \emph{what is shared} (pathology-relevant features that generalize across institutions) from \emph{what is source-specific} (acquisition and annotation artifacts), rather than leaving this separation to emerge from data alone. One could instead pursue domain-adversarial alignment~\cite{ganin2016domain} or naive joint training, but both are suboptimal: adversarial alignment risks removing clinically relevant variation (e.g., disease prevalence differences reflecting genuine epidemiological variation), while naive joint training encodes source-specific shortcuts 
that inflate in-domain scores at the cost of transferability. 
HarMoE takes a middle path: a shared decoder captures cross-dataset pathology semantics, augmented by lightweight low-rank residual modules \emph{deterministically routed by dataset identity}, reflecting that distributional shifts in multi-site imaging are 
systematic and institutional rather than instance-level. At inference, the residual modules are discarded; the training 
objective ensures the shared pathway alone carries sufficient diagnostic signal.

Class-level supervision sacrifices the compositional 
richness of free-text reports, but is more directly aligned with 
the zero-shot multi-label evaluation objective and, crucially, 
unlocks three datasets providing 496K additional images and 186 
additional disease categories inaccessible to any existing VLP 
method. The prompts retain sufficient semantic structure 
for the text encoder to generalize to unseen disease categories, 
as we demonstrate empirically.

\noindent Our contributions are as follows:

\noindent\textbf{(i)} We formalize multi-source chest X-ray pretraining under a unified disease vocabulary of 229 classes, constructed by harmonizing four major datasets (MIMIC-CXR, CheXpert, ChestX-ray14, PadChest) into a shared label space with tri-state encoding ($1$, $0$, $-1$ for present, absent, unknown) and masked supervision to prevent false negatives from unannotated entries.

\noindent\textbf{(ii)} We propose HarMoE, an architecture that 
separates shared pathology representations from dataset-specific 
residuals via constrained low-rank expert modules deterministically 
routed by dataset identity and discarded at inference, enabling 
multi-source scaling from 377K to 873K images without encoding 
source-dependent confounds into the transferable representation.

\noindent\textbf{(iii)} We evaluate on 11 benchmarks spanning in-domain and out-of-distribution settings and demonstrate consistent improvements in zero-shot classification, with the largest gains on out-of-distribution benchmarks: $+2.9\%$ AUC on RSNA, $+4.6\%$ AUC on COVID-QU-Ex, and $+4.1\%$ AUC on Montgomery. We additionally report improvements in zero-shot visual grounding without grounding-specific supervision.

\section{Related Work}

\noindent\textbf{Vision-Language Pretraining for Chest X-rays.}
Contrastive alignment between chest X-ray images and radiology reports has become the dominant pretraining paradigm. Early methods established global alignment (ConVIRT~\cite{zhang2022contrastive}) and local region-word matching (GLoRIA~\cite{huang2021gloria}), while BioViL~\cite{boecking2022making} and BioViL-T~\cite{bannur2023learning} extended this with token-level and temporal modeling. Knowledge-enhanced methods such as KAD~\cite{zhang2023knowledge} and MedKLIP~\cite{wu2023medklip} improve grounding through structured entity extraction. More recently, CarZero~\cite{lai2024carzero}, RadZero~\cite{park2025radzero}, and DeViDe~\cite{luo2024devide} move toward classification-level prompts or multi-attribute decomposition, while KEPIL~\cite{luo2026kepil} explicitly improves robustness to prompt variation through ontology-guided knowledge enrichment. MAVL~\cite{phan2024decomposing} decomposes visual features into attribute-specific components for finer-grained recognition. Complementary self-supervised approaches learn anatomy-aware chest X-ray representations through consistent embedding~\cite{zhou2023learning}, compositional decomposition~\cite{zhou2025ace}, or multi-perspective anatomical constraints~\cite{zhou2025lamps}. Beyond recognition, DWARF~\cite{luo2024dwarf} and hybrid explanation-guided learning~\cite{shu2025hybrid} refine diagnostic attention using expert or self-supervised guidance, while XBench~\cite{luo2026xbench} systematically evaluates visual-language explanations in chest radiography. These advances improve prompt robustness, anatomical representation, or grounding, but remain centered on single-source training. HarMoE instead exploits complementary supervision across heterogeneous classification-only datasets.

\noindent\textbf{Multi-Domain Learning and Dataset Bias in Medical Imaging.}
Distribution shifts across medical imaging datasets are well 
documented~\cite{oakden2020hidden, zech2018variable} and known to 
produce spurious correlations between source identity and diagnostic 
labels~\cite{degrave2021ai, geirhos2020shortcut}. Existing 
responses include domain-adversarial training~\cite{ganin2016domain}, 
invariant risk minimization~\cite{arjovsky2019invariant}, and 
adapter-based approaches such as LoRA~\cite{hu2022lora} and 
residual adapters~\cite{rebuffi2017learning}; partial-label 
learning~\cite{bekker2020learning} further addresses missing 
annotations in multi-label settings. Human-guided representation
alignment has likewise been shown to affect both out-of-distribution
generalization and demographic fairness in medical imaging~\cite{luo2025interplay}. HarMoE draws on these ideas
but makes a distinct choice: rather than adversarially removing 
domain information, which risks discarding clinically relevant 
variation, it deterministically assigns lightweight low-rank 
residual experts by dataset identity, confining source-specific 
variation to a controlled pathway that is discarded at inference, 
thereby preserving cross-dataset knowledge while preventing domain 
confounds from contaminating the transferable representation.

\noindent\textbf{Mixture-of-Experts Architectures.}
Sparse mixture-of-experts models, in which a learned router activates a subset of expert modules per input, have proven effective for scaling model capacity in language modeling~\cite{fedus2022switch} and vision~\cite{riquelme2021scaling}. In these systems, routing is input-conditioned and learned end-to-end. Our design departs from this convention: HarMoE routes deterministically by dataset identity, reflecting the fact that the dominant distributional shifts in multi-site medical imaging are systematic and institutional rather than instance-level. Each expert is a low-rank residual (LoRA) injected into the shared decoder, adding minimal parameters. This is closer in spirit to multi-domain adapters~\cite{rebuffi2017learning, hu2022lora} than to classical sparse MoE, and we use the term ``mixture of experts'' to denote the parallel expert bank and routing mechanism, not input-conditioned sparse gating.

\begin{figure*}[t]
    \centering
    \includegraphics[width=0.92\linewidth]{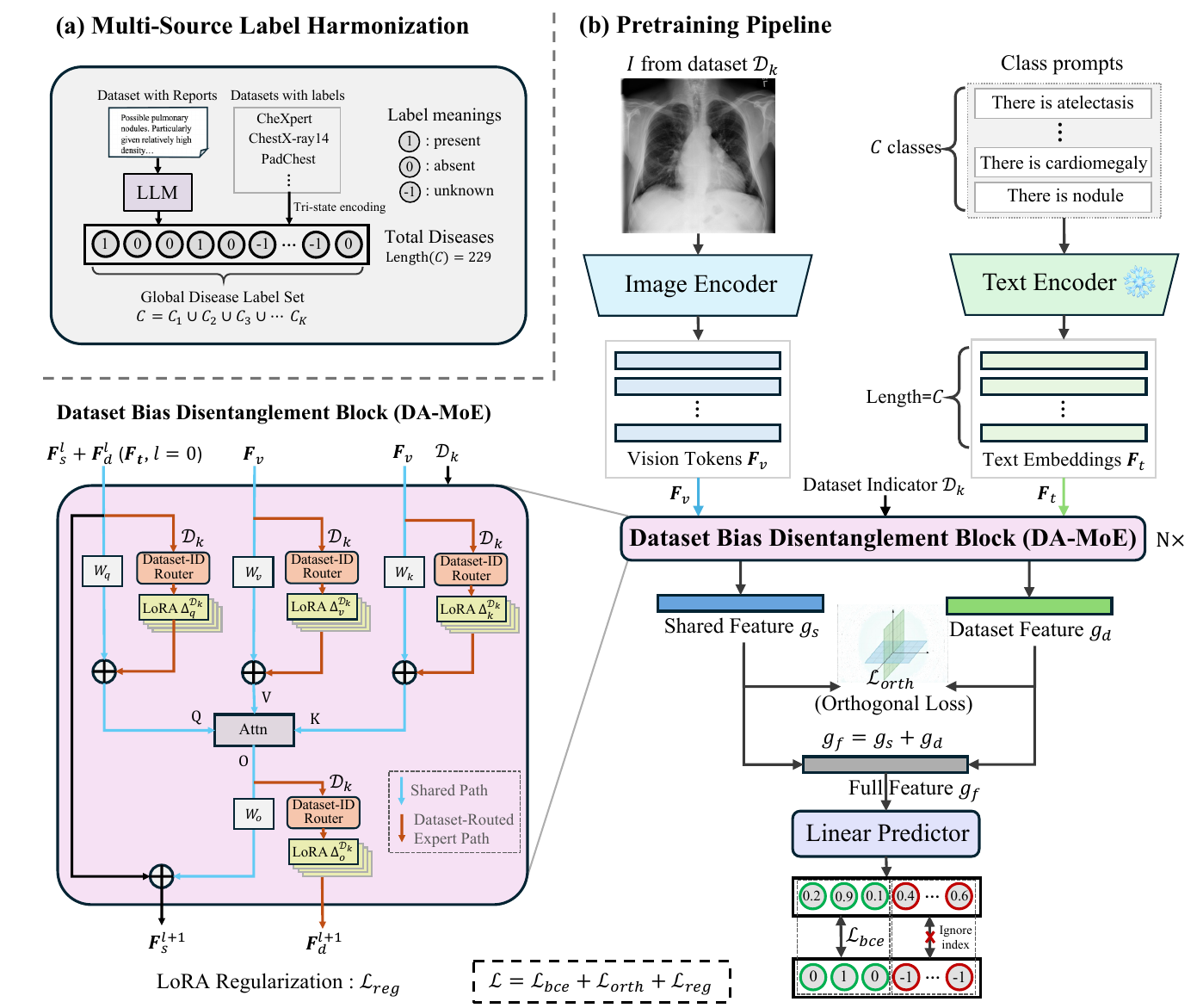}
    \Description{Pipeline diagram of label harmonization, shared and dataset-specific expert paths, masked training losses, and shared-only inference in HarMoE.}
    \caption{
    Overview of HarMoE.
    (a)~Annotations from $K$ datasets are unified into a 229-class
    vocabulary via LLM-based extraction and ontology alignment, producing
    tri-state labels ($1$/$0$/$-1$: present/absent/unknown).
    (b)~Disease prompts and visual tokens interact in $N$ decoder layers,
    each containing a shared path for dataset-invariant semantics and a
    dataset-routed expert path with low-rank (LoRA) residuals selected by
    dataset identity. The full representation
    $\mathbf{g}_f = \mathbf{g}_s + \mathbf{g}_d$ is used during training;
    at inference, the expert path is discarded and only $\mathbf{g}_s$
    drives prediction. The model is trained with masked classification
    ($\mathcal{L}_{\mathrm{bce}}$), orthogonality
    ($\mathcal{L}_{\mathrm{orth}}$), and expert regularization
    ($\mathcal{L}_{\mathrm{reg}}$).
     }
    \label{fig:benchmark_overview}
\end{figure*}


\section{Method}
\label{sec:method}

We present HarMoE, a framework for multi-source supervised pretraining over heterogeneous chest X-ray datasets with partial and inconsistent labels. The method has three components: (1)~\textbf{multi-source label harmonization} that unifies heterogeneous annotations into a shared disease vocabulary; (2)~\textbf{prompt-based multimodal encoding} that replaces free-text reports with stable disease-level text prompts, enabling the inclusion of classification-only datasets; and (3)~\textbf{dataset bias disentanglement block} (DA-MoE) that separates shared pathology representations from source-specific residuals via deterministic expert routing, with the experts discarded at inference to yield a domain-general representation. We describe each component below and conclude with the training objective.

\subsection{Multi-Source Label Harmonization}
\label{sec:unify_datasets}
The four pretraining datasets use incompatible annotation schemes: MIMIC-CXR~\cite{johnson2019mimic} provides free-text reports, CheXpert~\cite{irvin2019chexpert} uses a 14-class ontology with uncertainty labels, ChestX-ray14~\cite{wang2017chestx} provides 14 NLP-extracted binary labels, and PadChest~\cite{bustos2020padchest} annotates 174 findings. As shown in Fig. \ref{fig:benchmark_overview} (a), we harmonize multi-source datasets in three steps. First, for datasets with free-text reports, we apply an LLM to extract per-finding binary labels. Second, we construct a global vocabulary $\mathcal{C}$ of $|\mathcal{C}|=229$ conditions by taking the union of all dataset-specific label sets after resolving synonyms. Third, each sample is assigned a tri-state encoding label vector $\mathbf{y} \in \{-1, 0, 1\}^{|\mathcal{C}|}$, where $1$, $0$, and $-1$ denote present, absent, and unknown respectively; unknown entries are excluded from the loss rather than treated as negatives, preventing the false negatives when projecting sparse, dataset-specific annotations into a significantly expanded global vocabulary.

\subsection{Prompt-Based Multimodal Encoding}
\label{sec:encoding}
Building on the harmonized label space, we encode each disease concept as a fixed text embedding, bypassing the variance of free-text reports. As shown in Fig. \ref{fig:benchmark_overview} (b), for each condition $c \in \mathcal{C}$, we generate a prompt (``There is $c$.'') and encode it with a frozen clinical text encoder to obtain $\mathbf{F}_t \in \mathbb{R}^{|\mathcal{C}| \times D}$. Simultaneously, a vision backbone extracts visual tokens $\mathbf{F}_v \in \mathbb{R}^{P \times D}$ from the input radiograph, where $P$ and $D$ denote the number of patches and embedding dimension. This prompt-based interface provides a unified encoding for datasets with and without reports alike, and supports zero-shot generalization to unseen disease categories without architectural modification. 

\subsection{Dataset Bias Disentanglement}
\label{sec:decoder}

The central architectural contribution is a decoder that separates shared pathology representations from source-specific variation through two parallel paths: a \emph{shared path} that captures dataset-invariant cross-modal semantics, and a \emph{dataset-routed expert path} that absorbs source-dependent distributional differences via lightweight low-rank residual modules. The decoder consists of $N$ cascaded layers. At inference time, the expert path is discarded entirely, and only the shared representation is used for prediction. This is not a post-hoc simplification: the training objective (Section~\ref{sec:objective}) is specifically designed so that the shared path alone carries sufficient diagnostic signal, while the expert path serves as a controlled absorber of domain confounds during optimization.

\textbf{Shared Path.}
The shared path performs cross-attention between disease queries and visual tokens to distill pathology-relevant features that transfer across all sources. At the first layer ($l=1$), queries are derived from the text prompt embeddings $\mathbf{F}_t$ and cross-attend to visual tokens $\mathbf{F}_v$. For subsequent layers ($l > 1$), queries are updated from the previous shared output $\mathbf{F}_s^{(l-1)}$, while keys and values remain the visual tokens $\mathbf{F}_v$ throughout:
\begin{equation}
\mathbf{F}_s^{(l)} = \mathrm{Attn}\!\left(\mathbf{F}_s^{(l-1)} \mathbf{W}_q,\; \mathbf{F}_v \mathbf{W}_k,\; \mathbf{F}_v \mathbf{W}_v\right)\mathbf{W}_o,
\end{equation}
where $\mathbf{F}_s^{(0)} = \mathbf{F}_t$, and $\mathbf{W}_q, \mathbf{W}_k, \mathbf{W}_v, \mathbf{W}_o$ are learnable projection matrices shared across all datasets. This path is trained on data from all sources and captures the dataset-invariant component of the cross-modal alignment.

\textbf{Dataset-Routed Expert Path.}
In parallel, a dataset-routed path injects low-rank residual modules into the attention projections to model source-specific distributional differences. The design rationale is as follows: by modulating the query, key, value, and output projections of cross-attention, the expert modules can shift \emph{what the model attends to} and \emph{how it aggregates information} in a dataset-specific manner, without altering the shared feature space itself.
Specifically, a \textbf{dataset-ID router} receives the source indicator $\mathcal{D}_k$ and deterministically selects the corresponding expert from a bank of $\mathcal{K}$ experts. Routing is deterministic (not learned or input-conditioned) because the dominant distributional shifts across medical imaging datasets are systematic and institutional (scanner hardware, annotation conventions, patient demographics) rather than instance-level. For a given input $\mathbf{z}$ and attention projection $m \in \{q, k, v, o\}$, the expert-augmented projection is $\widetilde{\mathbf{z}}_m = \mathbf{z}\mathbf{W}_m + \alpha \cdot \mathbf{z}\mathbf{A}_m^{(k)}\mathbf{B}_m^{(k)}$,
where $\mathbf{W}_m$ is the shared projection (identical to the shared path), $\mathbf{A}_m^{(k)} \in \mathbb{R}^{D \times r}$ and $\mathbf{B}_m^{(k)} \in \mathbb{R}^{r \times D}$ are the low-rank expert parameters for dataset $k$, $r \ll D$ is the expert rank, and $\alpha$ controls the residual strength. Each expert adds only $2 \times 4 \times D \times r$ parameters per decoder layer, a small fraction of the shared decoder.
The query input mirrors the shared path: at layer $l$, queries take the dataset-routed decoder state $\mathbf{F}_d^{(l-1)}$ (with $\mathbf{F}_d^{(0)} = \mathbf{F}_t$), while keys and values take $\mathbf{F}_v$. The dataset-routed output at layer $l$ is:
\begin{equation}
\hat{\mathbf{F}}_d^{(l)} = \mathrm{Attn}\!\left(\widetilde{\mathbf{z}}^{(l)}_{q},\; \widetilde{\mathbf{z}}^{(l)}_k,\; \widetilde{\mathbf{z}}^{(l)}_v\right),
\end{equation}
\begin{equation}
\mathbf{F}_d^{(l)} = \hat{\mathbf{F}}_d^{(l)}\mathbf{W}_o + \alpha \cdot \hat{\mathbf{F}}_d^{(l)}\mathbf{A}_o^{(k)}\mathbf{B}_o^{(k)},
\end{equation}
where the output expert applies a final residual module. To allow the shared path to first establish a stable cross-modal alignment before introducing source-specific adaptation, expert modules are injected starting from the second decoder layer only.

\textbf{Additive Feature Decomposition.}
After $N$ decoder layers, the shared feature and dataset-routed output are summarized via mean pooling: 
$\mathbf{g}_s = \mathrm{MeanPool}(\mathbf{F}_s^{(N)})$, $\mathbf{g}_d = \mathrm{MeanPool}(\mathbf{F}_d^{(N)}).$\newline The full feature is their sum, $\mathbf{g}_f = \mathbf{g}_s + \mathbf{g}_d$.
This additive architecture explicitly decomposes the learned representation into a universal shared component and a source-specific correction. Consequently, $\mathbf{g}_s$ is optimized to encapsulate fundamental diagnostic signals independently of dataset-specific variances. Furthermore, the orthogonality constraint (Section~\ref{sec:objective}) ensures that $\mathbf{g}_s$ and $\mathbf{g}_d$ encode complementary, non-redundant information, preventing the loss of pathology-relevant features.

\textbf{Bias-Agnostic Inference Protocol.} At test time, the dataset-routed expert path is deactivated, allowing the model to rely solely on generalized pathology representations. The shared feature $\mathbf{g}_s$ alone is passed to the linear predictor to compute per-class diagnostic scores. This protocol effectively filters out imaging artifacts and institutional biases encoded in $\mathbf{g}_d$ without compromising prediction accuracy. Crucially, this bias-agnostic strategy requires no prior knowledge of a sample's origin, facilitating robust and seamless deployment across heterogeneous clinical environments.


\subsection{Training Objective}
\label{sec:objective}

The training objective combines three terms that ensure classification accuracy and effective bias disentanglement.

\textbf{Masked binary cross-entropy.}
A linear predictor maps the full feature $\mathbf{g}_f$ to per-class logits $\hat{y}_c$ for each disease $c \in \mathcal{C}$. Since each dataset annotates a different subset of conditions, we apply a masked loss that supervises only entries with known labels, directly implementing the tri-state encoding from Section~\ref{sec:unify_datasets}:
\begin{equation}
\mathcal{L}_{\mathrm{bce}} = \frac{1}{B}\sum_{i=1}^{B}\sum_{c\in\Omega_i}\mathrm{BCE}(\hat{y}_{i,c},\;y_{i,c}),
\end{equation}
where $\Omega_i \subseteq \mathcal{C}$ denotes the set of classes with known annotations ($y \in \{0,1\}$) for sample $i$ in a mini-batch of size $B$, and all unknown entries ($y=-1$) are excluded. This is the mechanism that prevents false negatives from unannotated classes: a 14-class dataset contributes supervision for its 14 conditions only, leaving the remaining 215 dimensions unaffected.

\textbf{Orthogonality loss.}
Without explicit constraints, the expert path may duplicate the shared representation, rendering the disentanglement trivial. We enforce complementarity by minimizing the squared cosine similarity between the shared and dataset-specific features:
\begin{equation}
\mathcal{L}_{\mathrm{orth}} = \frac{1}{B}\sum_{i=1}^{B}\left(\frac{\mathbf{g}_{s,i}^\top\,\mathbf{g}_{d,i}}{\|\mathbf{g}_{s,i}\|_2\,\|\mathbf{g}_{d,i}\|_2 + \epsilon}\right)^2.
\end{equation}
This encourages $\mathbf{g}_s$ and $\mathbf{g}_d$ to occupy orthogonal subspaces, ensuring that discarding $\mathbf{g}_d$ at inference removes source-specific variation without eliminating pathology-relevant signal.

\textbf{Expert regularization.}
Unconstrained low-rank experts may grow excessively large and overfit to source-specific artifacts. We penalize the Frobenius norm of the expert parameters:
\begin{equation}
\mathcal{L}_{\mathrm{reg}} = \frac{1}{N}\sum_{l=1}^{N}\sum_{m\in\{q,k,v,o\}}\left(\|\mathbf{A}_m^{(l)}\|_F^2 + \|\mathbf{B}_m^{(l)}\|_F^2\right),
\end{equation}
which keeps the expert residuals small relative to the shared projections, reinforcing their role as lightweight corrections rather than dominant feature extractors.

\textbf{Total objective.}
The three terms are combined as $\mathcal{L} = \mathcal{L}_{\mathrm{bce}} + \lambda_{\mathrm{orth}}\,\mathcal{L}_{\mathrm{orth}} + \lambda_{\mathrm{reg}}\,\mathcal{L}_{\mathrm{reg}}$,
where $\lambda_{\mathrm{orth}}$ and $\lambda_{\mathrm{reg}}$ control the strength of the decomposition constraints. 

\section{Experimental Settings}

\subsection{Pretraining Configuration}

\textbf{Pretraining datasets.} HarMoE is pretrained on the four-dataset consortium described in
Section~\ref{sec:unify_datasets}: MIMIC-CXR~\cite{johnson2019mimic}
(377K images), CheXpert~\cite{irvin2019chexpert} (224K images),
ChestX-ray14~\cite{wang2017chestx} (112K images), and
PadChest~\cite{bustos2020padchest} (160K images), totaling 873K
images over 229 unified disease classes. For datasets provided with unstructured reports (e.g., MIMIC-CXR), we utilize Qwen3~\cite{yang2025qwen3} as the large language model to distill structured tri-state labels. 

\textbf{Model Configurations.} The framework employs a ViT-B/16 as the image encoder, initialized with M3AE~\cite{chen2022multi} weights. For the text encoder, we utilize BioClinicalMPBERT~\cite{lai2024carzero} to encode disease prompts. The decoder consists of $N=4$ cascaded dataset bias disentanglement blocks. Dataset-routed expert paths are injected starting from the second layer. The LoRA expert rank is $r=8$, and the loss weights are set to $\lambda_{\mathrm{orth}} = \lambda_{\mathrm{reg}} = 1$.

\textbf{Optimization Strategy.} Training uses AdamW with an initial learning rate of $5 \times 10^{-5}$, weight decay of 0.02, and a cosine decay schedule over 100 epochs with a 20-epoch linear warm-up. The batch size is 32. The text encoder remains frozen throughout; the image encoder, decoder, and linear predictor are fully optimized. All experiments are conducted on two NVIDIA H200 GPUs with a model training time of approximately 72 hours.

\subsection{Evaluation Protocol}
\label{sec:eval_protocol}

\textbf{Zero-shot inference.}
All evaluations are performed without fine-tuning or linear probing.
The pretrained model is applied directly to each benchmark: for a given
test image, the shared feature $\mathbf{g}_s$ is extracted from the
decoder (with expert modules deactivated, as described in
Section~\ref{sec:decoder}) and passed through the linear predictor to
produce per-class scores.

\textbf{Benchmarks.}
We evaluate on 11 datasets spanning in-domain (ID) and out-of-distribution (OOD) settings. CheXpert and ChestX-ray14 serve as ID benchmarks because they overlap with the pretraining sources, although evaluation is conducted exclusively on their held-out test splits. The remaining benchmarks—OpenI~\cite{demner2016preparing}, RSNA~\cite{wu2024pneumonia}, ChestDR~\cite{wang2023real}, VinDr-CXR~\cite{nguyen2022vindr}, SIIM~\cite{siim-acr-pneumothorax-segmentation}, COVID-QU-Ex~\cite{tahir2021covid}, JSRT~\cite{shiraishi2000development}, Montgomery~\cite{jaeger2014two}, and Shenzhen~\cite{jaeger2014two}—are entirely unseen during pretraining and cover diverse acquisition conditions, patient populations, and disease categories, including COVID-19.

\textbf{Metrics.}
We report two complementary metrics. Area Under the ROC Curve (AUC)
measures ranking quality, the model's ability to assign higher scores to
positive cases. Matthews Correlation Coefficient (MCC) measures
class-level decision reliability at a fixed operating threshold,
capturing the balance between sensitivity and specificity. Reporting both
is important because a model with high AUC but low MCC produces good
rankings but poor binary decisions, a distinction that matters for
clinical deployment.

\textbf{Baselines.}
We compare against 11 vision-language methods spanning three categories: (i)~contrastive image-report alignment (ConVIRT~\cite{zhang2022contrastive}, GLoRIA~\cite{huang2021gloria}, BioViL~\cite{boecking2022making}, BioViL-T~\cite{bannur2023learning}, BioMedCLIP~\cite{zhang2023biomedclip}, CheXzero~\cite{tiu2022expert}); (ii)~knowledge-enhanced pretraining (KAD~\cite{zhang2023knowledge}, MedKLIP~\cite{wu2023medklip}, MAVL~\cite{phan2024decomposing}, DeViDe~\cite{luo2024devide}); and (iii)~prompt-level supervision (CarZero~\cite{lai2024carzero},  RadZero~\cite{park2025radzero}). All baselines are evaluated under the same zero-shot protocol for fair comparison.

\begin{table*}[t]
\centering
\setlength{\tabcolsep}{4pt}
\caption{Zero-shot multi-label classification (AUC, MCC) on two in-domain (ID) and three out-of-domain (OOD) benchmarks. Bold/underlined indicates best/second-best AUC. LT denotes datasets with long-tailed class distributions.}
\label{tab:multi-label}
\resizebox{\textwidth}{!}{%
\begin{tabular}{lcccccccccc}
\toprule
& \multicolumn{2}{c}{CheXpert (ID)}
& \multicolumn{2}{c}{ChestX-ray14 (ID)}
& \multicolumn{2}{c}{OpenI (OOD)}
& \multicolumn{2}{c}{ChestDR (OOD, LT)}
& \multicolumn{2}{c}{VinDr-CXR (OOD)} \\
\cmidrule(lr){2-3}\cmidrule(lr){4-5}\cmidrule(lr){6-7}
\cmidrule(lr){8-9}\cmidrule(lr){10-11}
Method
& AUC & MCC
& AUC & MCC
& AUC & MCC
& AUC & MCC
& AUC & MCC \\
\midrule
GLoRIA (ICCV 2021)
& 0.664 & 0.220
& 0.568 & 0.087
& 0.630 & 0.191
& 0.566 & 0.094
& 0.537 & 0.101 \\
ConVIRT (TMLR 2022)
& 0.807 & 0.396
& 0.651 & 0.132
& 0.637 & 0.188
& 0.664 & 0.165
& 0.669 & 0.170 \\
BioViL (ECCV 2022)
& 0.780 & 0.398
& 0.635 & 0.138
& 0.655 & 0.191
& 0.672 & 0.187
& 0.650 & 0.169 \\
CheXzero (Nature. Biomed. Eng. 2022)
& 0.890 & 0.554
& 0.685 & 0.159
& 0.703 & 0.255
& \underline{0.716} & \underline{0.220}
& 0.673 & 0.191 \\
BioViL-T (CVPR 2023)
& 0.827 & 0.451
& 0.631 & 0.136
& 0.662 & 0.175
& 0.714 & 0.211
& 0.664 & 0.180 \\

BioMedCLIP (NEJM AI 2023)
& 0.634 & 0.201
& 0.596 & 0.079
& 0.570 & 0.096
& 0.656 & 0.159
& 0.594 & 0.151 \\
KAD (Nature Com. 2023)
& 0.871 & 0.510
& 0.770 & 0.263
& 0.692 & 0.284
& 0.623 & 0.140
& 0.623 & 0.140 \\
MedKLIP (ICCV 2023)
& 0.911 & 0.581
& 0.727 & 0.201
& 0.588 & 0.138
& 0.630 & 0.146
& 0.604 & 0.160 \\
MAVL (CVPR 2024)
& 0.901 & \underline{0.606}
& 0.736 & 0.203
& 0.665 & 0.163
& 0.617 & 0.125
& 0.591 & 0.120 \\
CarZero (CVPR 2024)
& \underline{0.924} & \textbf{0.609}
& \underline{0.796} & 0.269
& 0.722 & 0.353
& 0.698 & 0.207
& 0.671 & 0.201 \\
DeViDe (IEEE BIBM 2025)
& 0.900 & 0.566
& 0.776 & \underline{0.270}
& 0.683 & 0.279
& 0.661 & 0.145
& 0.724 & \underline{0.274} \\
RadZero (NeurIPS 2025)
& 0.902 & 0.551
& 0.759 & 0.241
& \underline{0.818} & \underline{0.394}
& 0.704 & 0.213
& \underline{0.749} & \textbf{0.284} \\
\textbf{HarMoE (Ours)}
& \textbf{0.926}{ \color{red}(+0.2\%)} & 0.603
& \textbf{0.810}{ \color{red}(+1.4\%)} & \textbf{0.305}{ \color{red}(+3.5\%)}
& \textbf{0.836}{ \color{red}(+1.8\%)} & \textbf{0.464}{ \color{red}(+7\%)}
& \textbf{0.756}{ \color{red}(+4\%)} & \textbf{0.255}{\color{red}(+3.5\%)}
& \textbf{0.753}{ \color{red}(+0.4\%)} & 0.269 \\
\bottomrule
\end{tabular}}
\end{table*}

\begin{table*}[t]
\centering
\setlength{\tabcolsep}{4pt}
\caption{Zero-shot single-label classification (AUC, MCC) on six OOD benchmarks. Bold/underlined indicates best/second-best AUC.}
\label{tab:single-label}
\resizebox{\textwidth}{!}{%
\begin{tabular}{lcccccccccccc}
\toprule
& \multicolumn{2}{c}{RSNA (OOD)}
& \multicolumn{2}{c}{SIIM (OOD)}
& \multicolumn{2}{c}{COVID-QU-Ex (OOD)}
& \multicolumn{2}{c}{JSRT (OOD)}
& \multicolumn{2}{c}{Montgomery (OOD)}
& \multicolumn{2}{c}{Shenzhen (OOD)} \\
\cmidrule(lr){2-3}\cmidrule(lr){4-5}\cmidrule(lr){6-7}
\cmidrule(lr){8-9}\cmidrule(lr){10-11}\cmidrule(lr){12-13}
Method
& AUC & MCC
& AUC & MCC
& AUC & MCC
& AUC & MCC
& AUC & MCC
& AUC & MCC \\
\midrule
GLoRIA (ICCV 2021)
& 0.740 & 0.362
& 0.591 & 0.139
& 0.775 & 0.430
& 0.522 & 0.126
& 0.733 & 0.490
& 0.558 & 0.140 \\
ConVIRT (TMLR 2022)
& 0.576 & 0.167
& 0.630 & 0.221
& 0.681 & 0.393
& 0.649 & 0.272
& 0.842 & 0.738
& 0.816 & 0.529 \\
BioViL (ECCV 2022)
& 0.811 & 0.600
& 0.571 & 0.320
& 0.722 & 0.423
& 0.688 & 0.308
& 0.921 & 0.823
& 0.889 & 0.658 \\

CheXzero (Nature. Biomed. Eng. 2022)
& 0.900 & 0.636
& 0.784 & 0.393
& 0.517 & 0.057
& 0.653 & 0.315
& 0.784 & 0.504
& 0.675 & 0.311 \\

BioViL-T (CVPR 2023)
& 0.860 & 0.675
& 0.731 & 0.403
& 0.689 & 0.373
& 0.672 & 0.292
& 0.901 & 0.803
& 0.891 & 0.671 \\
BioMedCLIP (NEJM AI 2023)
& 0.778 & 0.414
& 0.629 & 0.201
& 0.337 & 0.000
& 0.481 & 0.126
& 0.725 & 0.571
& 0.802 & 0.474 \\
KAD (Nature Com. 2023)
& 0.778 & 0.414
& 0.874 & 0.561
& 0.728 & 0.446
& 0.607 & 0.205
& 0.897 & 0.783
& 0.804 & 0.558 \\
MedKLIP (ICCV 2023)
& 0.889 & 0.611
& 0.781 & 0.394
& 0.769 & 0.484
& 0.552 & 0.155
& 0.854 & 0.619
& 0.585 & 0.228 \\
MAVL (CVPR 2024)
& \underline{0.907} & \underline{0.647}
& 0.709 & 0.289
& \underline{0.870} & \underline{0.585}
& 0.624 & 0.267
& 0.845 & 0.692
& 0.650 & 0.317 \\
CarZero (CVPR 2024)
& 0.803 & 0.472
& 0.910 & 0.660
& 0.838 & 0.512
& 0.641 & 0.244
& \underline{0.909} & \underline{0.807}
& 0.773 & 0.532 \\
DeViDe (IEEE BIBM 2025)
& 0.887 & 0.625
& 0.895 & 0.609
& 0.730 & 0.470
& 0.658 & 0.271
& 0.879 & 0.742
& 0.871 & 0.586 \\
RadZero (NeurIPS 2025)
& 0.856 & 0.540
& \underline{0.931} & \underline{0.692}
& 0.832 & 0.562
& \underline{0.738} & \underline{0.367}
& 0.908 & 0.770
& \underline{0.916} & \underline{0.707} \\
\textbf{HarMoE (Ours)}
& \textbf{0.936}{ \color{red}(+2.9\%)} & \textbf{0.726}{ \color{red}(+7.9\%)}
& \textbf{0.943}{ \color{red}(+1.2\%)} & \textbf{0.719}{ \color{red}(+2.7\%)}
& \textbf{0.916}{ \color{red}(+4.6\%)} & \textbf{0.675}{ \color{red}(+9.0\%)}
& \textbf{0.740}{ \color{red}(+0.2\%)} & \textbf{0.423}{ \color{red}(+5.6\%)}
& \textbf{0.950}{ \color{red}(+4.1\%)} & \textbf{0.878}{ \color{red}(+7.1\%)}
& \textbf{0.925}{ \color{red}(+0.9\%)} & \textbf{0.738}{ \color{red}(+2.9\%)}\\
\bottomrule
\end{tabular}}
\end{table*}

\begin{table}[t]
\centering
\caption{Zero-shot visual grounding results (Dice score, Pointing game accuracy) on four benchmarks.}
\label{tab:dice_pointing_grid}
\resizebox{\linewidth}{!}{%
\begin{tabular}{lrrrrrrrrrr}
\toprule
\textbf{Model} & \multicolumn{2}{c}{\textbf{ChestXray14}} & \multicolumn{2}{c}{\textbf{RSNA}} & \multicolumn{2}{c}{\textbf{SIIM}} & \multicolumn{2}{c}{\textbf{Covid-Qu-Ex}} \\
 & \textbf{PG}$\uparrow$ & \textbf{Dice}$\uparrow$ 
 & \textbf{PG}$\uparrow$ & \textbf{Dice}$\uparrow$ 
 & \textbf{PG}$\uparrow$ & \textbf{Dice}$\uparrow$ 
 & \textbf{PG}$\uparrow$ & \textbf{Dice}$\uparrow$ \\
\cmidrule(lr){2-3}\cmidrule(lr){4-5}\cmidrule(lr){6-7}\cmidrule(lr){8-9}
\midrule

KAD & 0.350 & 0.236 & 0.701 & 0.421 & 0.010 & 0.043 & 0.120 & 0.315 \\

DeViDe & 0.372 & 0.247 & 0.708 & 0.402 & 0.065 & 0.069 & 0.074 & 0.325 \\

CARZero & 0.437 & 0.276 
& \underline{0.837} & \underline{0.505} 
& 0.135 & 0.112 
& 0.628 & 0.366 \\

RadZero & \underline{0.572} & \underline{0.392} 
& 0.765 & 0.446 
& \underline{0.210} & \textbf{0.163} 
& \textbf{0.688} & \textbf{0.458} \\

MAVL & 0.262 & 0.192 & 0.293 & 0.201 & 0.051 & 0.056 & 0.273 & 0.230 \\

MedKLIP & 0.318 & 0.233 & 0.428 & 0.331 & 0.010 & 0.037 & 0.408 & 0.299 \\

\textbf{HarMoE (Ours)} 
& \textbf{0.620} & \textbf{0.394} 
& \textbf{0.847} & \textbf{0.521} 
& \textbf{0.221} & \underline{0.156} 
& \underline{0.647} & \underline{0.422} \\

\bottomrule
\end{tabular}%
}
\end{table}

\section{Results}
\subsection{Zero-Shot Multi-Label Classification}
Tab.~\ref{tab:multi-label} reports zero-shot multi-label classification results on two in-domain benchmarks (CheXpert and ChestXray14) and three out-of-distribution benchmarks (OpenI, ChestDR, and VinDr-CXR). HarMoE achieves the best AUC on all five datasets, demonstrating that harmonized learning over heterogeneous supervision yields robust transfer beyond the pretraining domain. In particular, HarMoE improves over the second-best method by  +4.0\% on ChestDR, +1.4\% on ChestXray14, +1.8\% on OpenI, +0.2\% on CheXpert, and +0.4\% on VinDr-CXR in terms of AUC. The gains are especially pronounced on OOD datasets, where conventional VLP models are more easily affected by dataset-specific shortcut learning. HarMoE also delivers competitive or best MCC on most benchmarks, including clear improvements on ChestXray14 (+3.5\%), OpenI (+7.0\%), and ChestDR (+3.5\%), indicating that the performance gains are not limited to ranking quality but also translate into more reliable class-level predictions. Overall, these results suggest that unified multi-dataset supervision and controlled dataset-specific specialization enable the model to preserve transferable pathology semantics while reducing overfitting to individual source distributions.  

\subsection{Zero-Shot Single-Label Classification}  
HarMoE achieves state-of-the-art performance on all six OOD single-label benchmarks as shown in Tab.~\ref{tab:single-label}, with particularly striking gains on Montgomery (0.950 AUC, +4.1\%) and COVID-QU-Ex (0.916 AUC, +4.6\%), followed by RSNA (0.936 AUC, +2.9\%) and SIIM (0.943 AUC, +1.2\%). The corresponding MCC improvements are equally substantial: +7.9\% on RSNA, +9.0\% on COVID-QU-Ex, and +7.1\% on Montgomery, confirming that these gains reflect genuine class-level decision consistency. 
Notably, COVID-QU-Ex contains disease categories entirely absent from training data, yet HarMoE achieves its largest improvement there. This suggests the model leverages semantic correlations captured by the text encoder to infer unseen classes, rather than relying on dataset-specific visual shortcuts. These results validate the core HarMoE hypothesis: decoupling shared semantic knowledge from dataset-specific residuals through structured mixture-of-experts routing prevents overfitting to individual dataset statistics and yields harmonized representations that transfer robustly across heterogeneous medical domains.

\subsection{Zero-Shot Visual Grounding} Tab.~\ref{tab:dice_pointing_grid} reports zero-shot visual grounding results in terms of Dice score and pointing game accuracy. HarMoE achieves the highest scores on the majority of benchmarks, with the most notable improvement on ChestX-ray14, where pointing game accuracy rises from 0.572 to 0.620 (+4.8\% relative), followed by RSNA with Dice score improving from 0.505 to 0.521 and pointing game accuracy from 0.837 to 0.847. Importantly, no grounding-specific supervision is used during training; these gains arise solely from the cross-attention alignment learned under classification-level objectives.

\begin{figure*}[!t]
    \centering
    \includegraphics[width=\linewidth]{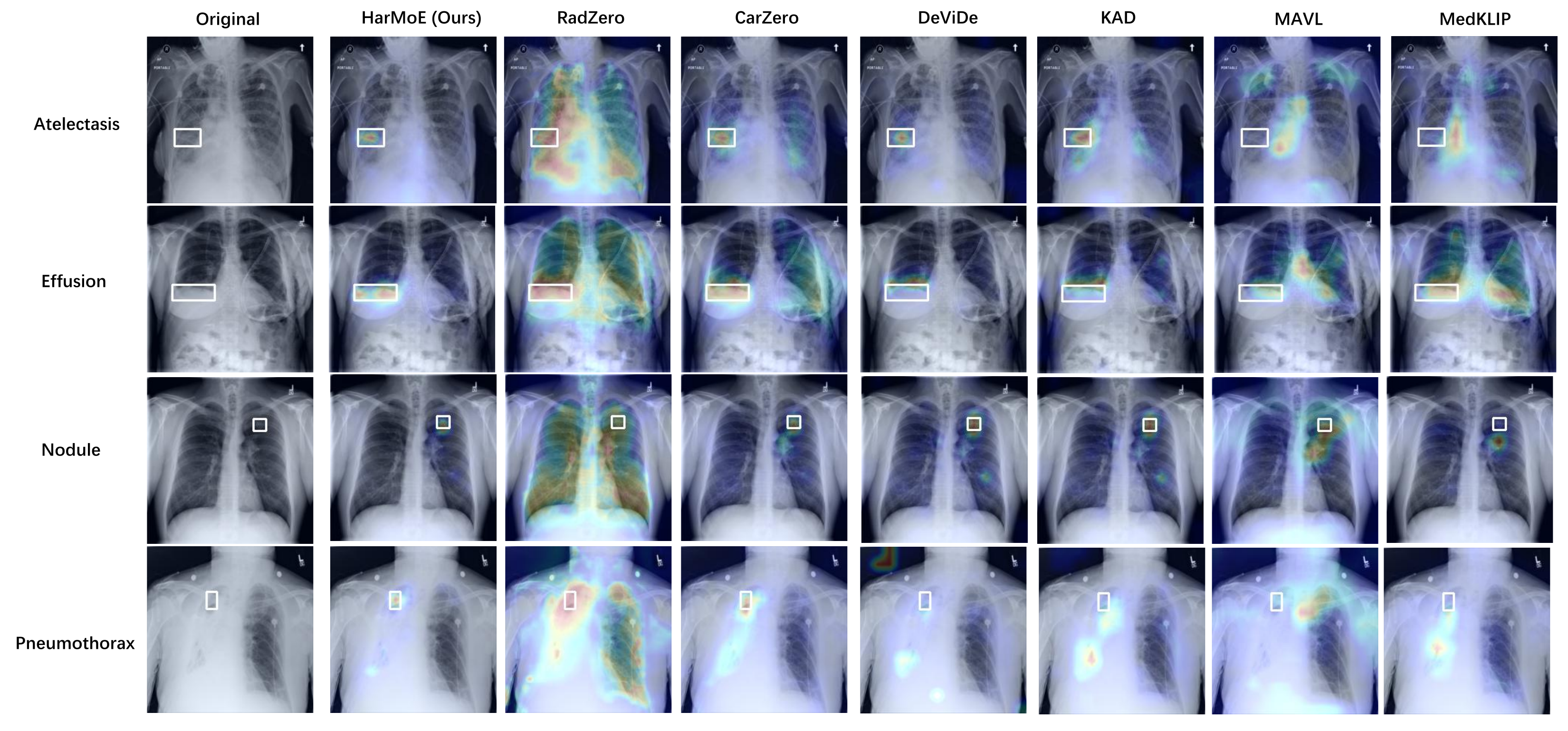}
    \Description{A grid of original radiographs and attention maps for four pathologies compares HarMoE with six prior methods.}
    \caption{
Qualitative comparison of zero-shot attention maps for four pathologies (Atelectasis, Effusion, Nodule, Pneumothorax) across methods. HarMoE produces more concentrated and anatomically aligned activations than prior methods.
    }
    \label{fig:attnmap}
\end{figure*}

Qualitative examples in Fig.~\ref{fig:attnmap} corroborate these quantitative findings. Prior VLP methods frequently produce diffuse or background-biased attention maps, whereas HarMoE yields spatially concentrated activations that align more closely with the anatomical extent of the target pathology. This effect is particularly evident for diffuse abnormalities (\emph{e.g.}, atelectasis, pleural effusion) and small focal lesions (\emph{e.g.}, nodule, pneumothorax). These results indicate that the shared--dataset feature decomposition introduced by HarMoE not only strengthens classification performance but also produces more spatially faithful cross-modal representations.

\begin{figure}[!t]
    \centering
    \includegraphics[width=0.9\columnwidth]{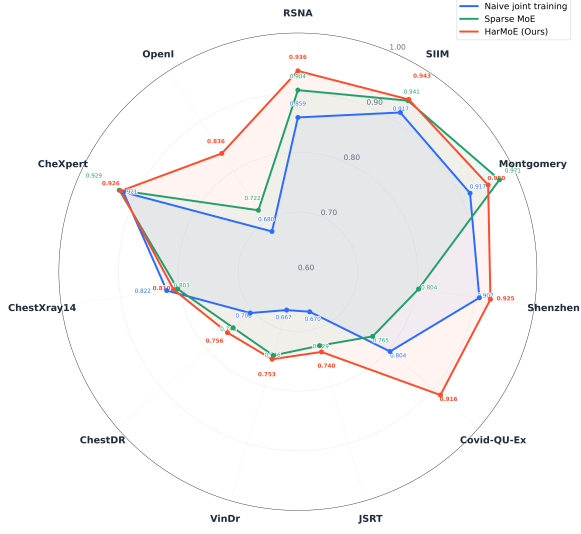}
    \Description{Radar chart comparing HarMoE, Sparse MoE, and naive joint training across in-domain and out-of-distribution evaluation benchmarks.}
    \caption{Radar plot comparing HarMoE, Sparse MoE, and naive joint training (w/o DA-MoE) across all evaluation benchmarks. DA-MoE improves out-of-distribution generalization by preventing the shared representation from overfitting to source-specific patterns 
    in the training data.}
    \label{fig:ablation_damoe_radar}
\end{figure}

\begin{figure}[t]
    \centering
    \includegraphics[width=\linewidth]{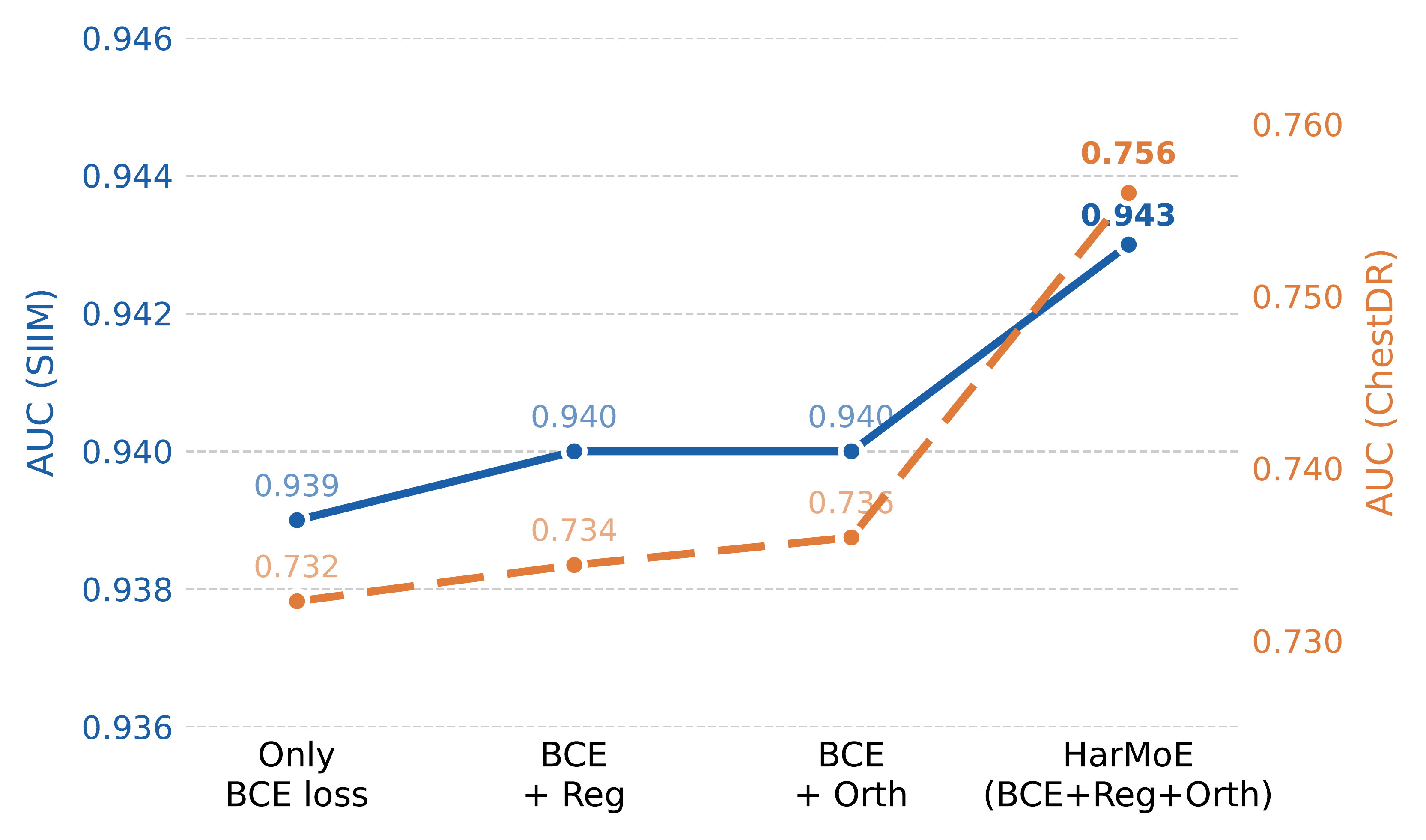}
    \Description{Bar-chart ablation comparing no regularization, residual regularization only, orthogonality regularization only, and both terms on SIIM and ChestDR.}
    \caption{Ablation on regularization terms (None, Res only, Orth only, Res+Orth) evaluated on SIIM and ChestDR. Combining both orthogonality and residual regularization yields the best performance.
    }
    \label{fig:ablation_orth}
\end{figure}

\begin{figure}[!t]
    \centering
    \includegraphics[width=0.9\columnwidth]{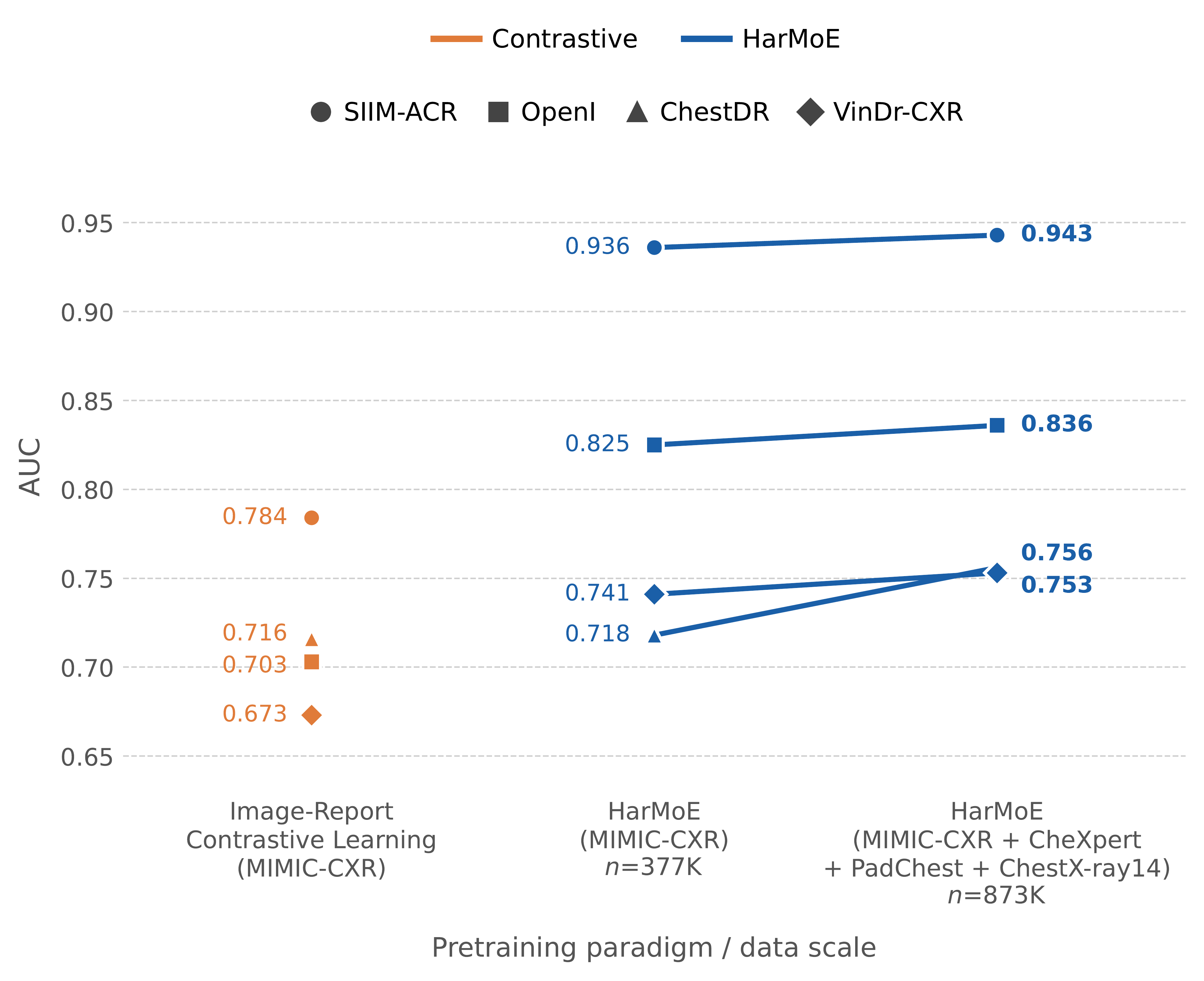}
    \Description{Line charts of AUC versus pretraining data scale for HarMoE and comparison paradigms across four out-of-distribution benchmarks.}
    \caption{Scaling behavior across pretraining paradigms and data scales on four OOD benchmarks. HarMoE with multi-source supervision scales more reliably than contrastive learning.}
    \label{fig:ablation_scale}
\end{figure}

\begin{figure}[!t]
    \centering
    \includegraphics[width=0.9\linewidth]{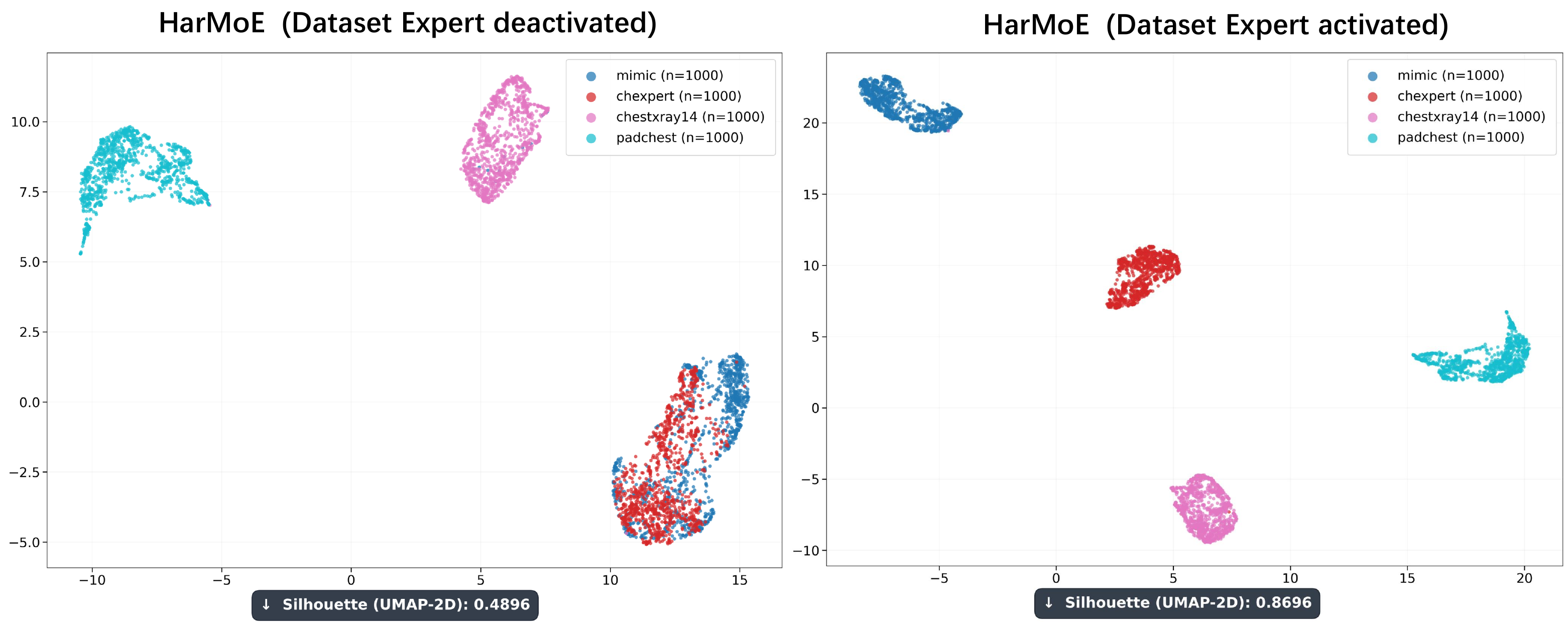}
    \Description{Two UMAP scatter plots show source-separated features with dataset experts active and aligned shared features with the experts inactive.}
    \caption{
    \textbf{UMAP visualization of learned representations.  Activating the dataset experts, features cluster by dataset source; deactivating the experts, the shared representation becomes better aligned across datasets.}
    }
    \label{fig:ablation_umap}
\end{figure}

\subsection{Ablation Study}

We conduct ablations to verify that HarMoE’s improvements stem from its architectural design choices. Specifically, we examine three aspects: (i)~the effect of the Dataset Bias Disentanglement Block (DA-MoE), (ii)~the contribution of individual regularization terms, and (iii)~the scaling behavior under multi-source supervision.

\textbf{Effect of DA-MoE.}
Fig.~\ref{fig:ablation_damoe_radar} compares HarMoE with Sparse MoE\cite{fedus2022switch} and naive joint training across all evaluation benchmarks. Naive joint training yields a modest improvement on the in-domain ChestX-ray14 benchmark (from 0.810 to 0.822 AUC) but causes consistent degradation on OOD benchmarks: Montgomery drops from 0.950 to 0.917 and VinDr-CXR from 0.753 to 0.667. This confirms that naive joint training without explicit shared--dataset decomposition encourages the model to encode source-specific shortcuts that inflate in-domain scores at the cost of transferability. While Sparse MoE partially recovers OOD performance over naive joint training,  HarMoE achieves consistently higher AUC across the most challenging cross-domain benchmarks, with notable improvements on OpenI (0.722 vs.\ 0.836) and Covid-QU-Ex (0.804 vs.\ 0.916), while remaining competitive on in-domain datasets such as ChestX-ray14 and CheXpert. This suggests that soft routing alone is insufficient to fully disentangle dataset-specific biases from transferable features, and that the explicit shared--private decomposition in DA-MoE is the key factor driving cross-domain generalization.
The UMAP visualization in Fig.~\ref{fig:ablation_umap} provides complementary evidence: activating the dataset expert, representations cluster primarily by dataset source, whereas deactivating the expert, the shared features become better aligned across datasets, with source-dependent variation confined to the residual branch.

\textbf{Effect of Regularization Terms.}
Fig.~\ref{fig:ablation_orth} isolates the contributions of the orthogonality loss $\mathcal{L}_{\mathrm{orth}}$ and the LoRA regularization $\mathcal{L}_{\mathrm{reg}}$. On SIIM, applying either term alone yields only marginal improvement over the unregularized baseline (0.940 \textit{vs.}\ 0.939), while their combination achieves 0.943. The effect is more pronounced on ChestDR, where neither term alone exceeds 0.736 AUC, but the full configuration reaches 0.756, a gain of 2.0\% absolute. These results indicate that the two terms play complementary roles: $\mathcal{L}_{\mathrm{orth}}$ prevents the residual branch from duplicating information already captured by the shared representation, while $\mathcal{L}_{\mathrm{reg}}$ constrains expert magnitude to preserve generalization. Their joint application is necessary for effective disentanglement.

\textbf{Effect of Scaling with Multi-Source Data.}
Fig.~\ref{fig:ablation_scale} compares three pretraining configurations on four OOD benchmarks: image--report contrastive learning on MIMIC-CXR, HarMoE trained on MIMIC-CXR alone, and HarMoE trained on all four source datasets. Two observations emerge. First, on the same data (MIMIC-CXR), replacing contrastive supervision with classification-level prompts yields consistent improvements across all benchmarks (\textit{e.g.}, SIIM AUC increases from 0.784 to 0.936). Second, incorporating additional heterogeneous datasets further improves all benchmarks ( ChestDR: 4\%, Vindr-CXR: 1.2\%), whereas contrastive learning offers no mechanism to incorporate classification-only datasets. These results demonstrate that HarMoE scales more effectively under multi-source supervision by leveraging cleaner supervisory signals while controlling dataset bias.


%


\section{Conclusion}
In this work, we introduced HarMoE, a dataset-specialized mixture-of-experts framework for learning from heterogeneous chest X-ray datasets. By incorporating classification-only datasets, HarMoE expands pretraining from 377K to 873K images while separating shared disease features from dataset-specific biases. Experiments show consistent improvements in zero-shot classification, out-of-distribution transfer, and visual grounding. These results demonstrate the value of structured multi-source supervision for building robust and generalizable radiology VLMs.

\bibliographystyle{ACM-Reference-Format}
\bibliography{samples/abbrev}

@STRING{jun = "June"}

@STRING{health = "ACM Transactions on Computing for Healthcare"}

@inproceedings{luo2024devide,
  title={DeViDe: Faceted Medical Knowledge to Enhance Vision Foundation Model Pretraining for Radiology},
  author={Luo, Haozhe and Zhou, Ziyu and Hou, Ming and Royer, Corentin and Reyes, Mauricio and Sekuboyina, Anjany and Menze, Bjoern},
  booktitle={2025 IEEE International Conference on Bioinformatics and Biomedicine (BIBM)},
  pages={1779--1782},
  publisher={IEEE},
  year={2025},
  doi={10.1109/BIBM66473.2025.11357035}
}

@article{luo2026kepil,
  title={{KEPIL}: Knowledge-Enhanced Prompt-Image Learning for Prompt-Robust Disease Detection},
  author={Luo, Haozhe and Shu, Shelley Zixin and Zhou, Ziyu and Berke, Robert and Reyes, Mauricio},
  journal={arXiv preprint arXiv:2605.09132},
  year={2026},
  doi={10.48550/arXiv.2605.09132}
}

@inproceedings{luo2026xbench,
  title={XBench: A Comprehensive Benchmark for Visual-Language Explanations in Chest Radiography},
  author={Luo, Haozhe and Shu, Shelley Zixin and Zhou, Ziyu and Ot{\'a}lora, Sebastian and Reyes, Mauricio},
  booktitle={2026 IEEE International Symposium on Biomedical Imaging (ISBI)},
  pages={1--5},
  publisher={IEEE},
  year={2026},
  doi={10.1109/ISBI61048.2026.11515566}
}

@inproceedings{zhou2025ace,
  title={{ACE}: Anatomically Consistent Embeddings in Composition and Decomposition},
  author={Zhou, Ziyu and Luo, Haozhe and Taher, Mohammad Reza Hosseinzadeh and Pang, Jiaxuan and Ding, Xiaowei and Gotway, Michael B. and Liang, Jianming},
  booktitle={2025 IEEE/CVF Winter Conference on Applications of Computer Vision (WACV)},
  pages={3823--3833},
  publisher={IEEE},
  year={2025},
  doi={10.1109/WACV61041.2025.00376}
}

@inproceedings{zhou2025lamps,
  title={Lamps: Learning Anatomy from Multiple Perspectives via Self-supervision in Chest Radiographs},
  author={Zhou, Ziyu and Luo, Haozhe and Taher, Mohammad Reza Hosseinzadeh and Pang, Jiaxuan and Ding, Xiaowei and Gotway, Michael B. and Liang, Jianming},
  booktitle={Foundation Models for General Medical AI},
  series={Lecture Notes in Computer Science},
  volume={16112},
  pages={1--11},
  publisher={Springer},
  year={2025},
  doi={10.1007/978-3-032-07845-2_1}
}

@inproceedings{luo2024dwarf,
  title={{DWARF}: Disease-Weighted Network for Attention Map Refinement},
  author={Luo, Haozhe and Pahud de Mortanges, Aur{\'e}lie and Inel, Oana and Reyes, Mauricio},
  booktitle={ISIC, iMIMIC, EARTH, and DeCaF Workshops at MICCAI},
  pages={59--68},
  publisher={Springer},
  year={2024},
  doi={10.1007/978-3-031-77610-6_6}
}

@inproceedings{shu2025hybrid,
  title={Hybrid Explanation-Guided Learning for Transformer-Based Chest X-Ray Diagnosis},
  author={Shu, Shelley Zixin and Luo, Haozhe and Poellinger, Alexander and Reyes, Mauricio},
  booktitle={Interpretability of Machine Intelligence in Medical Image Computing},
  series={Lecture Notes in Computer Science},
  volume={16464},
  pages={33--42},
  publisher={Springer},
  year={2025},
  doi={10.1007/978-3-032-17611-0_4}
}

@inproceedings{luo2025interplay,
  title={On the Interplay of Human-AI Alignment, Fairness, and Performance Trade-Offs in Medical Imaging},
  author={Luo, Haozhe and Zhou, Ziyu and Shu, Shelley Zixin and Pahud de Mortanges, Aur{\'e}lie and Berke, Robert and Reyes, Mauricio},
  booktitle={Medical Image Computing and Computer Assisted Intervention (MICCAI)},
  pages={420--430},
  publisher={Springer},
  year={2025},
  doi={10.1007/978-3-032-05185-1_41}
}

@inproceedings{lai2024carzero,
  title={Carzero: Cross-attention alignment for radiology zero-shot classification},
  author={Lai, Haoran and Yao, Qingsong and Jiang, Zihang and Wang, Rongsheng and He, Zhiyang and Tao, Xiaodong and Zhou, S Kevin},
  booktitle={Proceedings of the IEEE/CVF Conference on Computer Vision and Pattern Recognition},
  pages={11137--11146},
  year={2024}
}

@article{zhang2023knowledge,
  title={Knowledge-enhanced visual-language pre-training on chest radiology images},
  author={Zhang, Xiaoman and Wu, Chaoyi and Zhang, Ya and Xie, Weidi and Wang, Yanfeng},
  journal={Nature Communications},
  volume={14},
  number={1},
  pages={4542},
  year={2023},
  publisher={Nature Publishing Group UK London}
}

@inproceedings{radford2021learning,
  title={Learning transferable visual models from natural language supervision},
  author={Radford, Alec and Kim, Jong Wook and Hallacy, Chris and Ramesh, Aditya and Goh, Gabriel and Agarwal, Sandhini and Sastry, Girish and Askell, Amanda and Mishkin, Pamela and Clark, Jack and others},
  booktitle={International conference on machine learning},
  pages={8748--8763},
  year={2021},
  organization={PMLR}
}

@inproceedings{jia2021scaling,
  title={Scaling up visual and vision-language representation learning with noisy text supervision},
  author={Jia, Chao and Yang, Yinfei and Xia, Ye and Chen, Yi-Ting and Parekh, Zarana and Pham, Hieu and Le, Quoc and Sung, Yun-Hsuan and Li, Zhen and Duerig, Tom},
  booktitle={International conference on machine learning},
  pages={4904--4916},
  year={2021},
  organization={PMLR}
}

@inproceedings{bannur2023learning,
  title={Learning to exploit temporal structure for biomedical vision-language processing},
  author={Bannur, Shruthi and Hyland, Stephanie and Liu, Qianchu and Perez-Garcia, Fernando and Ilse, Maximilian and Castro, Daniel C and Boecking, Benedikt and Sharma, Harshita and Bouzid, Kenza and Thieme, Anja and others},
  booktitle={Proceedings of the IEEE/CVF Conference on Computer Vision and Pattern Recognition},
  pages={15016--15027},
  year={2023}
}

@article{shiraishi2000development,
  title={Development of a digital image database for chest radiographs with and without a lung nodule: receiver operating characteristic analysis of radiologists' detection of pulmonary nodules},
  author={Shiraishi, Junji and Katsuragawa, Shigehiko and Ikezoe, Junpei and Matsumoto, Tsuneo and Kobayashi, Takeshi and Komatsu, Ken-ichi and Matsui, Mitate and Fujita, Hiroshi and Kodera, Yoshie and Doi, Kunio},
  journal={American Journal of Roentgenology},
  volume={174},
  number={1},
  pages={71--74},
  year={2000},
  publisher={Am Roentgen Ray Soc}
}

@inproceedings{zhang2022contrastive,
  title={Contrastive learning of medical visual representations from paired images and text},
  author={Zhang, Yuhao and Jiang, Hang and Miura, Yasuhide and Manning, Christopher D and Langlotz, Curtis P},
  booktitle={Machine Learning for Healthcare Conference},
  pages={2--25},
  year={2022},
  organization={PMLR}
}

@inproceedings{huang2021gloria,
  title={Gloria: A multimodal global-local representation learning framework for label-efficient medical image recognition},
  author={Huang, Shih-Cheng and Shen, Liyue and Lungren, Matthew P and Yeung, Serena},
  booktitle={Proceedings of the IEEE/CVF International Conference on Computer Vision},
  pages={3942--3951},
  year={2021}
}

@inproceedings{boecking2022making,
  title={Making the most of text semantics to improve biomedical vision--language processing},
  author={Boecking, Benedikt and Usuyama, Naoto and Bannur, Shruthi and Castro, Daniel C and Schwaighofer, Anton and Hyland, Stephanie and Wetscherek, Maria and Naumann, Tristan and Nori, Aditya and Alvarez-Valle, Javier and others},
  booktitle={European conference on computer vision},
  pages={1--21},
  year={2022},
  organization={Springer}
}

@article{tiu2022expert,
  title={Expert-level detection of pathologies from unannotated chest X-ray images via self-supervised learning},
  author={Tiu, Ekin and Talius, Ellie and Patel, Pujan and Langlotz, Curtis P and Ng, Andrew Y and Rajpurkar, Pranav},
  journal={Nature Biomedical Engineering},
  volume={6},
  number={12},
  pages={1399--1406},
  year={2022},
  publisher={Nature Publishing Group UK London}
}

@article{johnson2019mimic,
  title={MIMIC-CXR-JPG, a large publicly available database of labeled chest radiographs},
  author={Johnson, Alistair EW and Pollard, Tom J and Greenbaum, Nathaniel R and Lungren, Matthew P and Deng, Chih-ying and Peng, Yifan and Lu, Zhiyong and Mark, Roger G and Berkowitz, Seth J and Horng, Steven},
  journal={arXiv preprint arXiv:1901.07042},
  year={2019}
}

@article{wu2024pneumonia,
  title={Pneumonia detection based on RSNA dataset and anchor-free deep learning detector},
  author={Wu, Linghua and Zhang, Jing and Wang, Yilin and Ding, Rong and Cao, Yueqin and Liu, Guiqin and Liufu, Changsheng and Xie, Baowei and Kang, Shanping and Liu, Rui and others},
  journal={Scientific Reports},
  volume={14},
  number={1},
  pages={1929},
  year={2024},
  publisher={Nature Publishing Group UK London}
}

@inproceedings{wang2017chestx,
  title={Chestx-ray8: Hospital-scale chest x-ray database and benchmarks on weakly-supervised classification and localization of common thorax diseases},
  author={Wang, Xiaosong and Peng, Yifan and Lu, Le and Lu, Zhiyong and Bagheri, Mohammadhadi and Summers, Ronald M},
  booktitle={Proceedings of the IEEE conference on computer vision and pattern recognition},
  pages={2097--2106},
  year={2017}
}

@misc{siim-acr-pneumothorax-segmentation,
    author = {Anna Zawacki, Carol Wu and George Shih, Julia Elliott and Mikhail Fomitchev, Mohannad Hussain, ParasLakhani and Phil Culliton, Shunxing Bao},
    title = {SIIM-ACR Pneumothorax Segmentation},
    publisher = {Kaggle},
    year = {2019},
    howpublished = {\url{https://kaggle.com/competitions/siim-acr-pneumothorax-segmentation}}
}

@inproceedings{irvin2019chexpert,
  title={Chexpert: A large chest radiograph dataset with uncertainty labels and expert comparison},
  author={Irvin, Jeremy and Rajpurkar, Pranav and Ko, Michael and Yu, Yifan and Ciurea-Ilcus, Silviana and Chute, Chris and Marklund, Henrik and Haghgoo, Behzad and Ball, Robyn and Shpanskaya, Katie and others},
  booktitle={Proceedings of the AAAI conference on artificial intelligence},
  volume={33},
  number={01},
  pages={590--597},
  year={2019}
}

@article{jaeger2014two,
  title={Two public chest X-ray datasets for computer-aided screening of pulmonary diseases},
  author={Jaeger, Stefan and Candemir, Sema and Antani, Sameer and W{\'a}ng, Y{\`\i}-Xi{\'a}ng J and Lu, Pu-Xuan and Thoma, George},
  journal={Quantitative imaging in medicine and surgery},
  volume={4},
  number={6},
  pages={475},
  year={2014},
  publisher={AME Publications}
}

@inproceedings{zhou2023learning,
  title={Learning Anatomically Consistent Embedding for Chest Radiography},
  author={Zhou, Ziyu and Luo, Haozhe and Pang, Jiaxuan and Ding, Xiaowei and Gotway, Michael and Liang, Jianming},
  booktitle={British Machine Vision Conference (BMVC)},
  pages={617--619},
  publisher={BMVA Press},
  year={2023},
  url={http://proceedings.bmvc2023.org/617/}
}

@article{bustos2020padchest,
  title={Padchest: A large chest x-ray image dataset with multi-label annotated reports},
  author={Bustos, Aurelia and Pertusa, Antonio and Salinas, Jose-Maria and De La Iglesia-Vaya, Maria},
  journal={Medical image analysis},
  volume={66},
  pages={101797},
  year={2020},
  publisher={Elsevier}
}

@inproceedings{phan2024decomposing,
  title={Decomposing disease descriptions for enhanced pathology detection: A multi-aspect vision-language pre-training framework},
  author={Phan, Vu Minh Hieu and Xie, Yutong and Qi, Yuankai and Liu, Lingqiao and Liu, Liyang and Zhang, Bowen and Liao, Zhibin and Wu, Qi and To, Minh-Son and Verjans, Johan W},
  booktitle={Proceedings of the IEEE/CVF Conference on Computer Vision and Pattern Recognition},
  pages={11492--11501},
  year={2024}
}

@inproceedings{wu2023medklip,
  title={Medklip: Medical knowledge enhanced language-image pre-training for x-ray diagnosis},
  author={Wu, Chaoyi and Zhang, Xiaoman and Zhang, Ya and Wang, Yanfeng and Xie, Weidi},
  booktitle={Proceedings of the IEEE/CVF International Conference on Computer Vision},
  pages={21372--21383},
  year={2023}
}

@article{zhang2023biomedclip,
  title={Biomedclip: a multimodal biomedical foundation model pretrained from fifteen million scientific image-text pairs},
  author={Zhang, Sheng and Xu, Yanbo and Usuyama, Naoto and Xu, Hanwen and Bagga, Jaspreet and Tinn, Robert and Preston, Sam and Rao, Rajesh and Wei, Mu and Valluri, Naveen and others},
  journal={arXiv preprint arXiv:2303.00915},
  year={2023}
}

@article{nguyen2022vindr,
  title={VinDr-CXR: An open dataset of chest X-rays with radiologist’s annotations},
  author={Nguyen, Ha Q and Lam, Khanh and Le, Linh T and Pham, Hieu H and Tran, Dat Q and Nguyen, Dung B and Le, Dung D and Pham, Chi M and Tong, Hang TT and Dinh, Diep H and others},
  journal={Scientific Data},
  volume={9},
  number={1},
  pages={429},
  year={2022},
  publisher={Nature Publishing Group UK London}
}

@article{park2025radzero,
  title={RadZero: Similarity-Based Cross-Attention for Explainable Vision-Language Alignment in Radiology with Zero-Shot Multi-Task Capability},
  author={Park, Jonggwon and Kim, Soobum and Yoon, Byungmu and Choi, Kyoyun},
  journal={arXiv e-prints},
  pages={arXiv--2504},
  year={2025}
}

@article{gundel2021robust,
  title={Robust classification from noisy labels: Integrating additional knowledge for chest radiography abnormality assessment},
  author={G{\"u}ndel, Sebastian and Setio, Arnaud AA and Ghesu, Florin C and Grbic, Sasa and Georgescu, Bogdan and Maier, Andreas and Comaniciu, Dorin},
  journal={Medical Image Analysis},
  volume={72},
  pages={102087},
  year={2021},
  publisher={Elsevier}
}

@article{clark1995interobserver,
  title={Interobserver Variability in Interpreting Chest Radiographs},
  author={Clark, Dwayne C and Conrad, KA},
  journal={Archives of Internal Medicine},
  volume={155},
  number={13},
  pages={1453--1453},
  year={1995},
  publisher={American Medical Association}
}

@article{bekker2020learning,
  title={Learning from positive and unlabeled data: A survey},
  author={Bekker, Jessa and Davis, Jesse},
  journal={Machine learning},
  volume={109},
  number={4},
  pages={719--760},
  year={2020},
  publisher={Springer}
}

@article{geirhos2020shortcut,
  title={Shortcut learning in deep neural networks},
  author={Geirhos, Robert and Jacobsen, J{\"o}rn-Henrik and Michaelis, Claudio and Zemel, Richard and Brendel, Wieland and Bethge, Matthias and Wichmann, Felix A},
  journal={Nature Machine Intelligence},
  volume={2},
  number={11},
  pages={665--673},
  year={2020},
  publisher={Nature Publishing Group UK London}
}

@inproceedings{oakden2020hidden,
  title={Hidden stratification causes clinically meaningful failures in machine learning for medical imaging},
  author={Oakden-Rayner, Luke and Dunnmon, Jared and Carneiro, Gustavo and R{\'e}, Christopher},
  booktitle={Proceedings of the ACM conference on health, inference, and learning},
  pages={151--159},
  year={2020}
}

@article{zech2018variable,
  title={Variable generalization performance of a deep learning model to detect pneumonia in chest radiographs: a cross-sectional study},
  author={Zech, John R and Badgeley, Marcus A and Liu, Manway and Costa, Anthony B and Titano, Joseph J and Oermann, Eric Karl},
  journal={PLoS medicine},
  volume={15},
  number={11},
  pages={e1002683},
  year={2018},
  publisher={Public Library of Science}
}

@article{demner2016preparing,
  title={Preparing a collection of radiology examinations for distribution and retrieval},
  author={Demner-Fushman, Dina and Kohli, Marc D and Rosenman, Marc B and Shooshan, Sonya E and Rodriguez, Laritza and Antani, Sameer and Thoma, George R and McDonald, Clement J},
  journal={Journal of the American Medical Informatics Association},
  volume={23},
  number={2},
  pages={304--310},
  year={2016},
  publisher={Oxford University Press}
}

@article{wang2023real,
  title={A real-world dataset and benchmark for foundation model adaptation in medical image classification},
  author={Wang, Dequan and Wang, Xiaosong and Wang, Lilong and Li, Mengzhang and Da, Qian and Liu, Xiaoqiang and Gao, Xiangyu and Shen, Jun and He, Junjun and Shen, Tian and others},
  journal={Scientific Data},
  volume={10},
  number={1},
  pages={574},
  year={2023},
  publisher={Nature Publishing Group UK London}
}

@article{tahir2021covid,
  title={COVID-19 infection localization and severity grading from chest X-ray images},
  author={Tahir, Anas M and Chowdhury, Muhammad EH and Khandakar, Amith and Rahman, Tawsifur and Qiblawey, Yazan and Khurshid, Uzair and Kiranyaz, Serkan and Ibtehaz, Nabil and Rahman, M Sohel and Al-Maadeed, Somaya and others},
  journal={Computers in biology and medicine},
  volume={139},
  pages={105002},
  year={2021},
  publisher={Elsevier}
}

@inproceedings{cherti2023reproducible,
  title={Reproducible scaling laws for contrastive language-image learning},
  author={Cherti, Mehdi and Beaumont, Romain and Wightman, Ross and Wortsman, Mitchell and Ilharco, Gabriel and Gordon, Cade and Schuhmann, Christoph and Schmidt, Ludwig and Jitsev, Jenia},
  booktitle={Proceedings of the IEEE/CVF conference on computer vision and pattern recognition},
  pages={2818--2829},
  year={2023}
}

@article{degrave2021ai,
  title={AI for radiographic COVID-19 detection selects shortcuts over signal},
  author={DeGrave, Alex J and Janizek, Joseph D and Lee, Su-In},
  journal={Nature Machine Intelligence},
  volume={3},
  number={7},
  pages={610--619},
  year={2021},
  publisher={Nature Publishing Group UK London}
}

@article{fedus2022switch,
  title={Switch transformers: Scaling to trillion parameter models with simple and efficient sparsity},
  author={Fedus, William and Zoph, Barret and Shazeer, Noam},
  journal={Journal of Machine Learning Research},
  volume={23},
  number={120},
  pages={1--39},
  year={2022}
}

@inproceedings{chen2022multi,
  title={Multi-modal masked autoencoders for medical vision-and-language pre-training},
  author={Chen, Zhihong and Du, Yuhao and Hu, Jinpeng and Liu, Yang and Li, Guanbin and Wan, Xiang and Chang, Tsung-Hui},
  booktitle={International Conference on Medical Image Computing and Computer-Assisted Intervention},
  pages={679--689},
  year={2022},
  organization={Springer}
}

@inproceedings{ganin2016domain,
  title={Domain-Adversarial Training of Neural Networks},
  author={Ganin, Yaroslav and Ustinova, Evgeniya and Ajber, Hana and Germain, Pascal and Larochelle, Hugo and Laviolette, Fran{\c{c}}ois and Marchand, Mario and Lempitsky, Victor},
  booktitle={Journal of Machine Learning Research},
  volume={17},
  number={59},
  pages={1--35},
  year={2016}
}

@inproceedings{arjovsky2019invariant,
  title={Invariant Risk Minimization},
  author={Arjovsky, Martin and Bottou, L{\'e}on and Gulcevich, Ishaan and Lopez-Paz, David},
  booktitle={arXiv preprint arXiv:1907.02893},
  year={2019}
}

@inproceedings{hu2022lora,
  title={{LoRA}: Low-Rank Adaptation of Large Language Models},
  author={Hu, Edward J. and Shen, Yelong and Wallis, Phillip and Allen-Zhu, Zeyuan and Li, Yuanzhi and Wang, Shean and Wang, Lu and Chen, Weizhu},
  booktitle={International Conference on Learning Representations (ICLR)},
  year={2022}
}

@inproceedings{rebuffi2017learning,
  title={Learning Multiple Visual Domains with Residual Adapters},
  author={Rebuffi, Sylvestre-Alvise and Bilen, Hakan and Vedaldi, Andrea},
  booktitle={Advances in Neural Information Processing Systems (NeurIPS)},
  volume={30},
  year={2017}
}

@inproceedings{riquelme2021scaling,
  title={Scaling Vision with Sparse Mixture of Experts},
  author={Riquelme, Carlos and Puigcerver, Joan and Mustafa, Basil and Neumann, Maxim and Jenatton, Rodolphe and Susano Pinto, Andr{\'e} and Keysers, Daniel and Houlsby, Neil},
  booktitle={Advances in Neural Information Processing Systems (NeurIPS)},
  volume={34},
  year={2021}
}

@article{yang2025qwen3,
  title={Qwen3 technical report},
  author={Yang, An and Li, Anfeng and Yang, Baosong and Zhang, Beichen and Hui, Binyuan and Zheng, Bo and Yu, Bowen and Gao, Chang and Huang, Chengen and Lv, Chenxu and others},
  journal={arXiv preprint arXiv:2505.09388},
  year={2025}
}

\end{document}